\documentclass[letterpaper, 10 pt, conference]{ieeeconf}
\IEEEoverridecommandlockouts    
\title{\LARGE \bf
AERMANI-Diffusion: Regime-Conditioned Diffusion for Dynamics Learning in Aerial Manipulators 
}

\author{Samaksh Ujjawal$^{*1}$, Shivansh Pratap Singh$^{*1}$, Naveen Sudheer Nair$^{1}$, \\ Rishabh Dev Yadav$^{2}$,Wei Pan$^{2}$, Spandan Roy$^{1}$
\thanks{$^{1}$ Robotics Research Center, International Institute of Information Technology Hyderabad, Hyderabad {\tt \footnotesize (\{samaksh.ujjawal, shivansh.singh\}@research.iiit.ac.in, \ spandan.roy@iiit.ac.in}).  }
\thanks{$^{2}$ Department of Computer Science, The University of Manchaster, U.K. ({\tt \footnotesize  rishabh.yadav@postgrad.manchaster.ac.uk, \  wei.pan@pmanchaster.ac.uk).} }
\thanks{$^{*}$ Equal contribution. }
}
\usepackage{amsmath}
\usepackage{amssymb}
\usepackage{mathtools}
\usepackage{algorithm, algpseudocode}
\usepackage{graphicx, wrapfig, xcolor, cite}
\usepackage{url}
\usepackage{multirow}

\usepackage[utf8]{inputenc}
\usepackage{float}
\usepackage{bm}          % For bold math symbols
\usepackage{caption}
\usepackage{booktabs}    % For professional tables
\usepackage{cite}        % For bundled citations [1-3]
\usepackage{cuted}       % For full-page-width equations if needed
\usepackage{lipsum}      % (For placeholder text, remove later)

\usepackage{bbm}

\newcommand{\norm}[1]{\left\lVert#1\right\rVert}

\newcommand{\tbr}[1]{\left[#1\right]}

\newcommand{\expectation}[2]{\mathop{\mathbb{E}}_{#2}\tbr{#1}}

\newtheorem{theorem}{Theorem}[section]
\newtheorem{lemma}[theorem]{Lemma}

\newtheorem{remark}{Remark}

\newcommand{\bz}{\mathbf{z}}

\DeclareMathOperator{\diag}{diag}

\usepackage{array}
\algnewcommand{\Initialise}[1]{%
  \State \textbf{Initialise:}
  \Statex \hspace*{\algorithmicindent}\parbox[t]{.8\linewidth}{\raggedright #1}
}

\usepackage{xcolor}
\begin{document}

\maketitle
\thispagestyle{empty}
\pagestyle{empty}

\begin{abstract}   
Aerial manipulators undergo rapid, configuration-dependent changes in inertial coupling forces and aerodynamic forces, making accurate dynamics modeling a core challenge for reliable control. 
Analytical models lose fidelity under these nonlinear and nonstationary effects, while standard data-driven methods such as deep neural networks and Gaussian processes cannot represent the diverse residual behaviors that arise across different operating conditions.
We propose a regime-conditioned diffusion framework that models the full distribution of residual forces using a conditional diffusion process and a lightweight temporal encoder. The encoder extracts a compact summary of recent motion and configuration, enabling consistent residual predictions even through abrupt transitions or unseen payloads.
When combined with an adaptive controller, the framework enables dynamics uncertainty compensation and yields markedly improved tracking accuracy in real-world tests.
\end{abstract}

% \begin{keyword}
% { Aerial Manipulation; Learning-Based Control; Diffusion Models; Data-Driven Modeling; Adaptive Control}
% \end{keyword}

%===============================================================================

\section{Introduction}

Aerial manipulators combine a quadrotor platform with articulated robotic arms, enabling tasks such as inspection, contact-based manipulation, and payload transport in complex 3D workspaces (\cite{10339889, 10758214}). Unlike standalone quadrotors, these systems experience strong and rapidly varying dynamic coupling between the aerial platform and the manipulator stemming from manipulator motion, payload changes, and, from residual forces generated by shifts in the center of mass and inertia 
(\cite{suarez2020benchmarks, 9813358}). Achieving high-precision control in such conditions requires reliable compensation of these residual effects. However, obtaining their accurate analytical models is very difficult, if at all possible, due to nonlinear, state-dependent interactions and unpredictable aerodynamic effects such as downwash, rotor–manipulator interference etc. (\cite{8059875, ruggiero2018aerial, meng2020survey}). 

\begin{figure}[!h]
\centering
\includegraphics[width=0.98\linewidth]{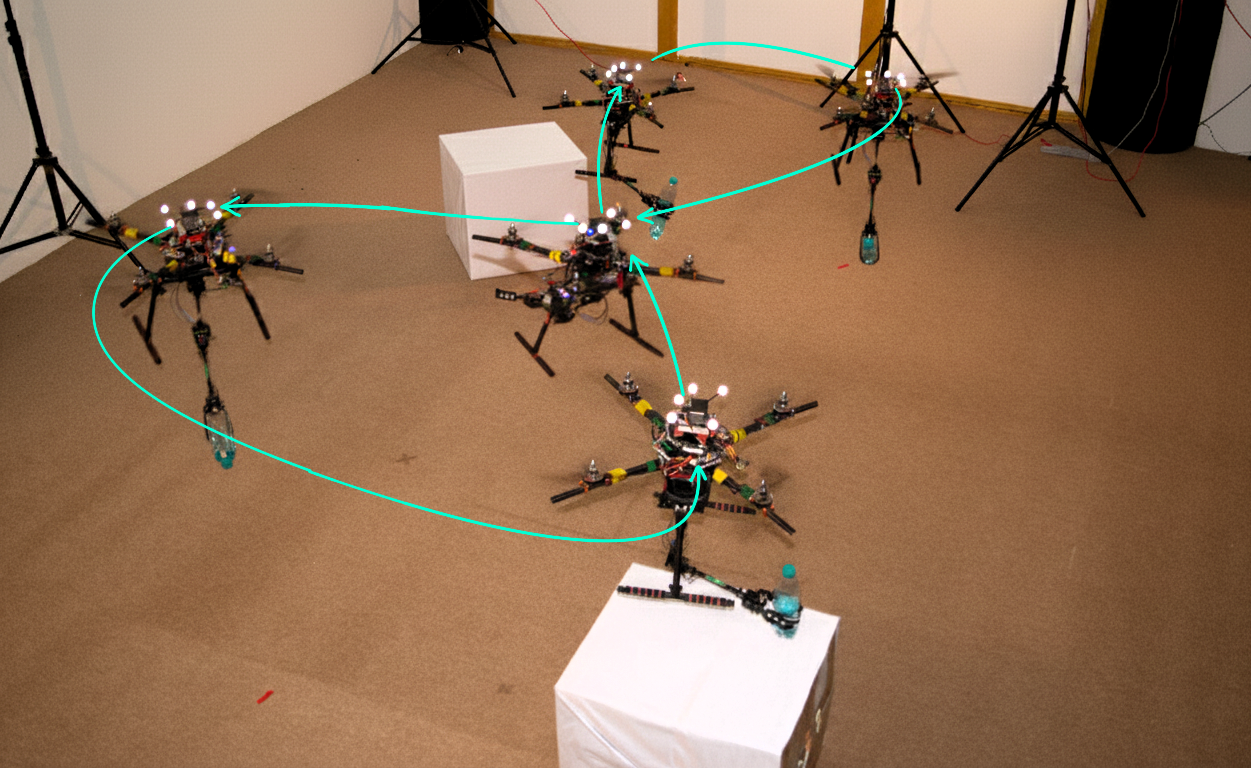}
% \vspace{-2mm}
\caption{\small Real-world experiment showing aerial manipulator executing a dynamic 3D trajectory while carrying a payload. } \label{fig:real_exp_trajectory}
\end{figure}

\subsection{Related Works and Contribution}
% \vspace{-2mm}
Figure \ref{fig:real_exp_trajectory} shows a popular pick-and-place operation by an aerial manipulator, which consists of three distinct operational phases: free flight, grasping and payload transport, and payload release. The transitions from free flight to grasping and transportation to payload release are particularly characterized by non-smooth changes in inertial and interaction forces due to sudden changes in system mass; whereas, the movement of manipulator during transportation gives rise to dynamic changes in center of mass.
Since no single model can remain accurate across these phases, conventional control methods struggle to compensate for large variations in inertia (\cite{9462539}) and configuration-dependent coupling forces between the vehicle and manipulator (\cite{orsag2017dexterous}). While a range of data-driven models have been found effective for quadrotors, they face certain fundamental limitations when applied to model an aerial manipulator (\cite{saviolo2023learning}). We briefly discuss them in the following.

Deep Neural Networks (DNNs) (\cite{li2017deep, shi2019neural,salzmann2023real}) learn a single deterministic mapping from inputs to outputs. This forces them to average over uncertainty, ignore variability in residual forces, and produce predictions with no calibrated confidence, making them unreliable outside the training set. Gaussian Processes (GPs) (\cite{torrente2021data, crocetti2023gapt, cao2024computation}) scale poorly with dataset size and rely on kernels that impose smooth, stationary residual dynamics. These assumptions limit their ability to model transient nonlinear interactions, and non-Gaussian disturbances that commonly arise in aerial manipulation.
NeuralODEs (\cite{chee2022knode}) and NeuralSDEs (\cite{chee2022knode}) assume smooth continuous-time dynamics, which fails in aerial manipulators where residual forces contain nonsmooth effects from contact and abrupt inertial or aerodynamic changes.
Physics-informed Temporal Convolutional Networks (TCNs) (\cite{saviolo2022physics}) improve temporal feature extraction but still encode a single nominal dynamics pattern, which cannot represent multiple qualitatively different behaviors in an aerial manipulator. 

Aerial manipulator dynamics and uncertainties are influenced by latent effects such as airflow interactions, downwash, and internal coupling forces, which behave as hidden states that onboard sensing cannot capture~(\cite{Bauersfeld2021NeuroBEMHA}). Sequence models, including RNN-based predictors~(\cite{mohajerin2019multistep}), long-horizon multi-step models~(\cite{rao2024learning}), and transformer architectures~(\cite{chen2021decision}), leverage temporal structure and can, in principle, encode information about operating conditions. However, they still return a single deterministic prediction with no notion of variability, and therefore, cannot account for the different operational regimes of an aerial manipulator. Their autoregressive nature also leads to error accumulation, limiting their effectiveness during fast transitions. 
Recent diffusion-based approaches~(\cite{das2025dronediffusion}) improve robustness by learning a full residual distribution, but they condition only on instantaneous measurements and thus cannot infer which operating condition produced them. This reveals a key gap: how to infer the operating regime in a principled way while learning the system’s dynamics.

To tackle the modeling challenges of aerial manipulators, we propose a conditional diffusion model augmented with a temporal encoding of recent motion. This combination provides reliable residual predictions without relying on detailed dynamics modeling and integrates cleanly with an adaptive controller for stable flight. The resulting framework improves prediction fidelity and tracking robustness without imposing restrictive assumptions on the underlying dynamics. The highlights of this work are enumerated below: % contributions are:

\begin{itemize}
\item \textbf{A formal residual-dynamics formulation for aerial manipulators.}
We show that the interaction forces arising from manipulator motion, payload variation, and configuration changes cannot be represented by a single smooth mapping, and we explicitly characterize them as a sequence-conditioned prediction problem with regime-dependent structure. This provides a principled foundation for learning-based compensation without requiring acceleration measurements or detailed inertia matrix.

\item \textbf{A regime-conditioned diffusion model for residual prediction.}
We introduce a conditional diffusion framework in which a lightweight temporal encoder extracts a compact regime descriptor from a short history window. This enables the model to separate behavior across operating conditions, generate consistent residual estimates, and remain stable during rapid configuration changes.

\item \textbf{Real-world validation on an aerial manipulator.} We integrate the learned residual model into an adaptive controller and obtain robust tracking across payload changes, fast manipulator motions, and out-of-distribution conditions. Experiments show consistent improvements in predictive accuracy and closed-loop performance, with the controller compensating effectively for both learned and unmodeled dynamics.
\end{itemize}

\section{Methodology}
We denote $t$ as the physical time index, $S$ as the prediction horizon, and $k$ as the diffusion timestep; subscript $(\cdot)_t$ refers to a physical-time quantity at $t$, whereas superscript $(\cdot)^k$ refers to its noisy value at diffusion step $k$.

\subsection{Preliminaries on Diffusion Models}
\label{sec:preliminaries}

Denoising Diffusion Probabilistic Models (DDPMs)~(\cite{ho2020denoising}) function as parameterized Markov chains that learn to generate data matching a target distribution $q(\bz^0)$ by reversing a fixed forward diffusion process. The forward process, $q(\bz^k|\bz^{k-1}) \coloneqq \mathcal{N}(\bz^k|\sqrt{1-\beta^k}\bz^{k-1}, \beta^k\mathbf{I})$, systematically destroys structure in the data by injecting Gaussian noise over $K$ timesteps, governed by a variance schedule $\{\beta^k\}_{k=1}^K$, such that $\bz^K$ approximates an isotropic Gaussian $\mathcal{N}(\mathbf{0}, \mathbf{I})$.

The generative reverse process, $p_{\theta}(\bz^{k-1}|\bz^k)$, learns to denoise the latent variables to recover the original data structure. Training optimizes a variational lower bound on the log-likelihood, which simplifies to a score-matching objective:
\begin{equation}
\label{eqn:ho_simple}
\mathcal{J}_{\rm denoise}(\theta)\coloneqq \hspace{-10pt} \expectation{\norm{\epsilon - \epsilon_\theta(\bz^k, k)}^2}{k\sim\mathcal{U}(1, K), \bz^0 \sim q, \epsilon \sim \mathcal{N}(\mathbf{0}, \mathbf{I})},
\end{equation}
where $\epsilon_\theta$ is a neural network parameterized by $\theta$ that predicts the noise component $\epsilon$. To enable control applications, the reverse process is conditioned on an observation context $c$, modeling the conditional distribution $p_\theta(\bz^0|c)$ via $\epsilon_\theta(\bz^k, c, k)$.
\subsection{Residual Dynamics as Sequence Generation}
\label{sec:dynamics_formulation}

We consider a quadrotor-based aerial manipulator with $n$ degrees of freedom manipulator and generalized coordinates $\chi \in \mathbb{R}^{6+n}$, driven by control inputs $\tau \in \mathbb{R}^{6+n}$. Its motion is governed by the Euler--Lagrange dynamics (\cite{arleo2013control})
\begin{equation}
\label{eq:EL}
    M(\chi)\ddot{\chi} + C(\chi,\dot{\chi})\dot{\chi} + g(\chi) + d(\dot{\chi}, t) = \tau,
\end{equation}
where $M(\chi)$ is the inertia matrix, $C(\chi,\dot{\chi})$ contains Coriolis/centripetal effects, $g(\chi)$ is gravity, and $d(\dot{\chi}, t)$ represents unmodeled aerodynamic and external disturbances.  
% For aerial manipulators, these dynamics exhibit nonstationary behavior that is difficult to represent by fixed parametric models. 
Further, the inertia matrix admits the standard block decomposition
\begin{align}
\label{split}
M = 
\begin{bmatrix}
    M_{pp} & M_{p\alpha} \\
    M_{p\alpha}^{\top} & M_{\alpha\alpha}
\end{bmatrix}, \qquad
\begin{matrix}
M_{pp}\!\in\!\mathbb{R}^{6\times6},\;
M_{p\alpha}\!\in\!\mathbb{R}^{6\times n},\\[1mm]
M_{\alpha\alpha}\!\in\!\mathbb{R}^{n\times n},
\end{matrix}
\end{align}
where the off–diagonal block $M_{p\alpha}$ captures the inertial coupling between the quadrotor and the manipulator. These couplings depend on manipulator configuration and introduce nonlinear, state-dependent interaction forces that are difficult to characterize reliably using analytic expressions \cite[Ch.~5.3]{orsag2018aerial}. As a result, model-based controllers relying on imperfect coupling models often suffer from degraded accuracy or instability when the system configuration or payload changes.

To isolate these unknown effects, we introduce a user-defined nominal inertia matrix $\bar M$ and rewrite \eqref{eq:EL} as
\begin{equation}
\label{eq:residual_def}
    \bar M\,\ddot{\chi} + \mathcal{H}(\chi,\dot{\chi},\ddot{\chi}, t) = \tau,
\end{equation}
with
$ \mathcal{H} \triangleq
    [M(\chi)-\bar M]\ddot{\chi}
    + C(\chi,\dot{\chi})\dot{\chi}
    + g(\chi)
    + d(\dot{\chi}, t). $
The term $\mathcal{H}$ consolidates all nonstationary, unmodeled, configuration-dependent, and disturbance contributions into a single residual vector. This reformulation shifts the dynamics-learning problem from estimating each physical component individually to directly learning $\mathcal{H}$, which is more practical for aerial manipulator whose dynamics vary with task conditions, manipulator posture, or payload.

Since $\mathcal{H}$ evolves along the trajectory and depends on the recent motion history, treating it as an independent pointwise regression target leads to temporally inconsistent predictions and error accumulation. Instead, we model the residual dynamics as a sequence-generation problem. For a prediction horizon $S$, we construct overlapping trajectory segments of the form
\begin{align}
\mathcal{S}_t
=
\{\zeta_{t:t+S-1},\;
  \tau_{t:t+S-1},\;
  \mathcal{H}_{t:t+S-1}\},
 ~\zeta_t \triangleq (\chi_t,\dot{\chi}_t),
\end{align}
which record the evolution of states, inputs, and residuals over $S$ consecutive time steps.  
These fixed-length segments serve as the fundamental training samples from which the diffusion model learns the relationship between residual forces and the corresponding state–input sequences.

\begin{figure}[!h]
\centering
\includegraphics[scale=0.275]{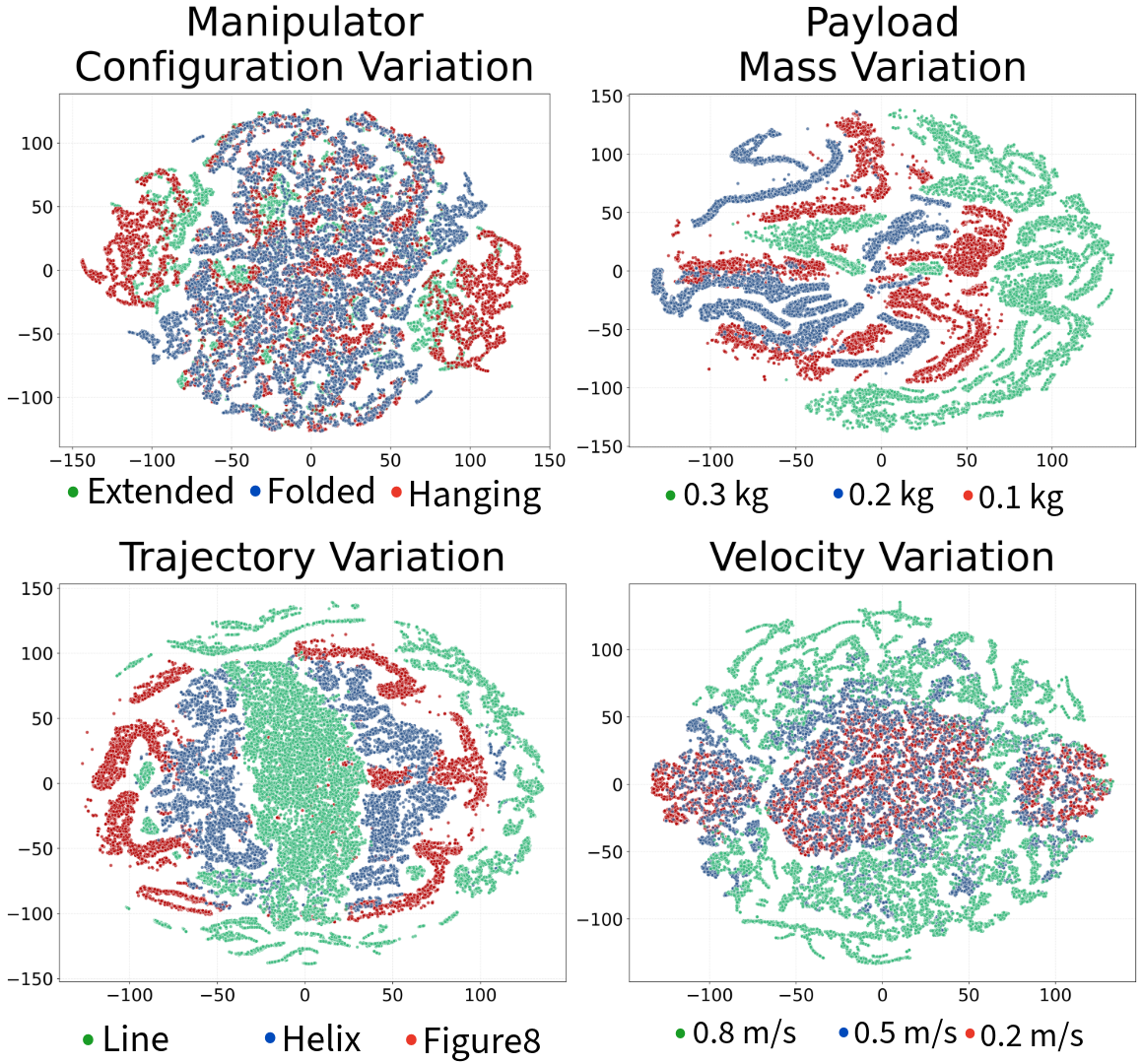}   

    \caption{\small  t-SNE embedding of the residual dynamics $\mathcal{H}$ collected from real-world flights. 
Each panel isolates a single factor of variation: (top-left) manipulator configuration, (top-right)  payload mass, (bottom-left) trajectory geometry, and (bottom-right) flight velocity.  Distinct clusters emerge for each condition, forming curved and partially overlapping  manifolds. These structures suggests the regime-dependent nature of the  residual dynamics.}
    \label{fig:regimes}

\end{figure}

\subsection{Regime-Dependent Residual Dynamics}
\label{sec:regime_structure}

The residual forces acting on an aerial manipulator depend strongly on operating conditions such as manipulator configuration, payload mass, and recent motion. To study this behavior, we project the measured residuals $\mathcal{H}$ onto a two-dimensional t-SNE embedding using real flight data collected under diverse trajectories and payloads. Figure~\ref{fig:regimes} reveals three consistent patterns.

First, the residuals group into distinct clusters associated with different operating conditions (e.g., payload on/off, manipulator posture), indicating that the mapping $(\zeta,\tau)\!\mapsto\!\mathcal{H}$ changes across regimes. Second, within each group the residuals vary smoothly, forming continuous curves rather than isolated points, reflecting natural variations due to motion and velocity changes. Third, the clusters partially overlap, meaning that the same instantaneous state--input pair $(\zeta,\tau)$ may produce different residual forces depending on the underlying operating condition or recent motion history.

This behavior is characteristic of systems whose dynamics change with configuration or payload. Although the aerial manipulator evolves continuously, variations in manipulator posture or payload effectively switch the underlying dynamics, producing regime-specific behaviors that overlap in the state--input space. Consequently, the instantaneous pair $(\zeta,\tau)$ does not uniquely determine the resulting residual force.

We formalize this by defining \emph{regime-dependent residual dynamics}:  the residual $\mathcal{H}$ is said to be regime dependent if the conditional distribution of $S$  given the instantaneous state--input pair $(\zeta, \tau)$ contains multiple distinct clusters, each corresponding to different operating conditions.  Since the underlying regime is not directly observable from $(\zeta, \tau)$ alone, the resulting conditional distribution appears multi-clustered, 
as illustrated in Figure~\ref{fig:regimes}.

\subsection{Regime-Conditioned Diffusion Model}
\label{sec:regime_conditioned_diffusion}

As shown in Section~\ref{sec:regime_structure}, the residual learning cannot rely on a single deterministic map. Instead, an accurate model must represent the full conditional distribution of $\mathcal{H}$ across regimes. Diffusion models support this requirement by learning distributions with multiple separated clusters without enforcing restrictive smoothness assumptions, motivating their use for residual dynamics in aerial manipulation.

To capture these regime-dependent effects, we introduce a lightweight encoder that extracts a compact descriptor from a short history of recent motion:
\begin{equation}
    r_t = f_\phi(\zeta_{t-L:t}, \tau_{t-L:t}).
\end{equation}
The history window provides information that is not contained in a single measurement, such as payload engagement, configuration-induced inertia variations, and aerodynamic changes. It also conveys higher-order motion trends without relying on noisy numerical differentiation of $\dot{\chi}$. We implement the encoder as a shallow temporal convolutional network (TCN), which is sufficient to capture these short-term signatures while keeping computation low for real-time deployment.

\begin{figure}[!h]
\centering
\includegraphics[width=0.98\linewidth]{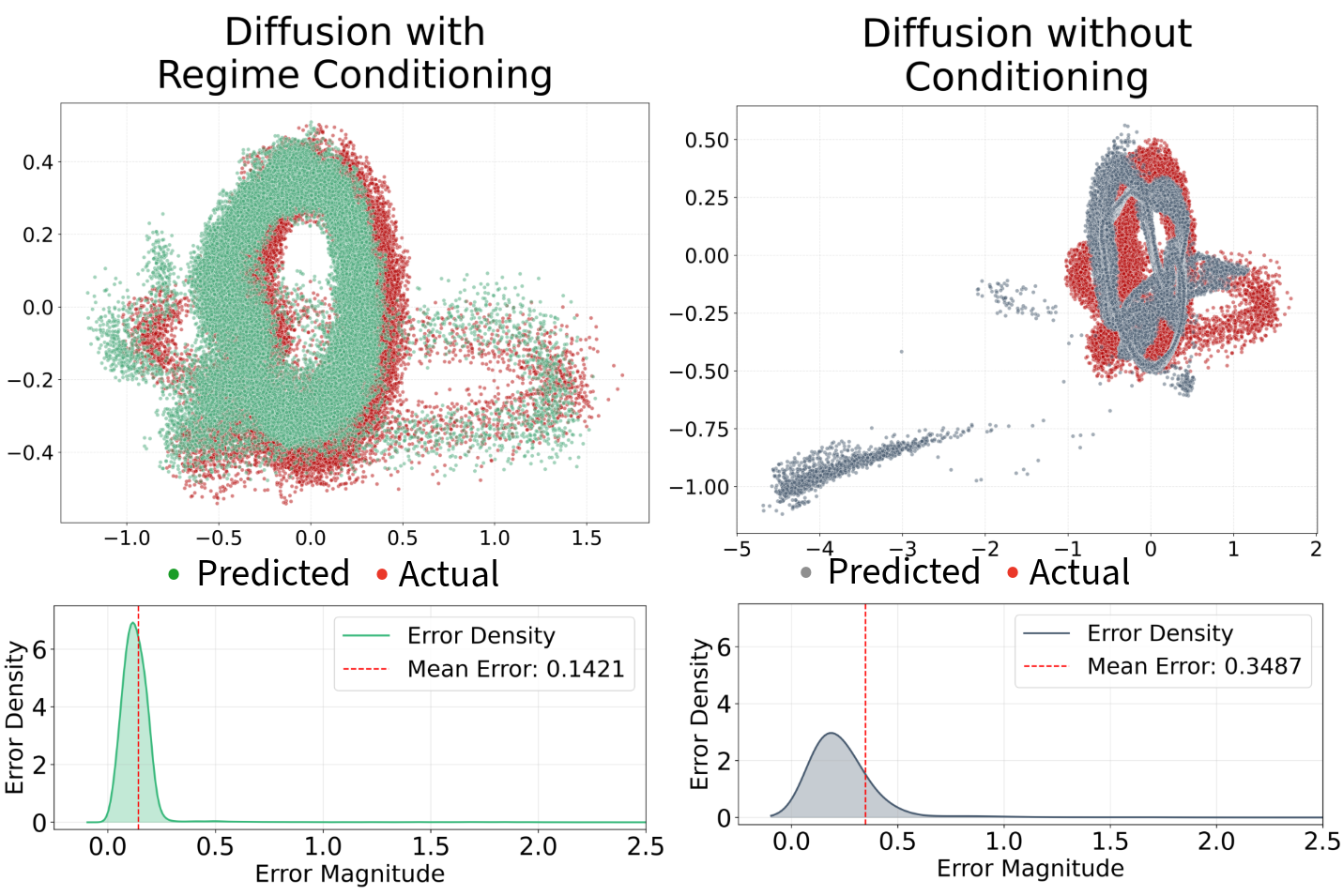}
    \caption{\small 
Comparison of diffusion models with and without regime conditioning. 
Top: 2D PCA projection of predicted residual dynamics $\hat{\mathcal{H}}$.
Bottom: corresponding error distributions.}
\label{fig:regime_ablation}
\end{figure}

We model the residual dynamics using Denoising Diffusion Probabilistic Models (DDPMs)~(\cite{ho2020denoising}), whose reverse process is interpreted through Stochastic Langevin Dynamics~(\cite{welling2011bayesian}). 
Following conditional diffusion formulations used in robotics (\cite{chi2023diffusionpolicy}), we learn the conditional distribution $p(\mathcal{H} \mid \zeta, \tau, r)$, where $\mathcal{H}^{k}$ denotes the noisy residual at diffusion step $k$ in the reverse denoising process.
 The reverse update is
\begin{equation}
\label{eq:reverse_regime}
    \mathcal{H}^{k-1}
        \hspace{-3pt} = \hspace{-3pt}\frac{1}{\sqrt{\alpha^{k}}}
          \left(
              \mathcal{H}^{k}
              \hspace{-3pt} - \hspace{-3pt} \frac{1-\alpha^{k}}{\sqrt{1-\bar{\alpha}^{k}}}
                \,\epsilon_{\theta} (\zeta,\tau,r,\mathcal{H}^{k},k)
          \right)
          + \sqrt{\beta^{k}}\,z,
\end{equation}
where $\alpha^{k}=1-\beta^{k}$, $\bar{\alpha}^{k}=\prod_{i=1}^{k}\alpha^{i}$, $z\sim\mathcal{N}(0,I)$, and $\epsilon_{\theta}$ is the denoiser and training is carried out using an offline dataset $\mathcal{D}=\{\mathcal{S}_i\}_{i=1}^{N}$ of state, input, and residual sequences. The denoising process
\eqref{eq:reverse_regime}, the simplified objective \eqref{eqn:ho_simple} becomes:
\begin{equation}
\label{eq:ddpm_regime_loss_explicit}
\mathcal{J}(\theta,\phi)
=
\mathbb{E}_{\substack{
k\sim\mathcal{U}(1,K),\;
\epsilon\sim\mathcal{N}(0,I),\\
(\zeta,\tau,\mathcal{H}^0)\sim\mathcal{D}
}}
\Big[
\|\epsilon - \epsilon_{\theta}(\zeta,\tau,r,\mathcal{H}^{k},k)\|^{2}
\Big],
\end{equation}
with gradients flowing through both the diffusion model parameters $\theta$ and the encoder parameters $\phi$. No auxiliary objectives or separate training stages are required. At deployment, a rolling buffer maintains the most recent $L{+}1$ observations to compute $r_t$. The reverse process~\eqref{eq:reverse_regime} then generates a sample $\hat{\mathcal{H}}$, and the first-step output $\hat{\mathcal{H}}^{1}$ is supplied to the model-based controller. 

As shown in Figure \ref{fig:regime_ablation}, conditioning on the regime descriptor prevents the diffusion model from mixing incompatible behaviors, yielding residual predictions that closely follow the true dynamics. Without this conditioning, the model collapses trajectories across different operating conditions and produces large, inconsistent errors.

\begin{remark}
Conditioning the diffusion model on the full state–action history would, in principle, supply the temporal context that a regime encoder provides. In practice, however, this history forms a high-dimensional and redundant signal that is difficult to learn from limited data and is known to destabilize score estimation~\cite{2025ldp}. Although the model is trained on residual \emph{sequences}, its denoiser receives only the \emph{current} state and input at inference, so it cannot recover temporal cues on its own. A compact latent descriptor $r_t$, extracted from a short history window, provides this information efficiently and yields stable conditioning without the cost of full-history inputs.
\end{remark}

\begin{figure}[!h]
\centering
\includegraphics[width=0.98\linewidth]{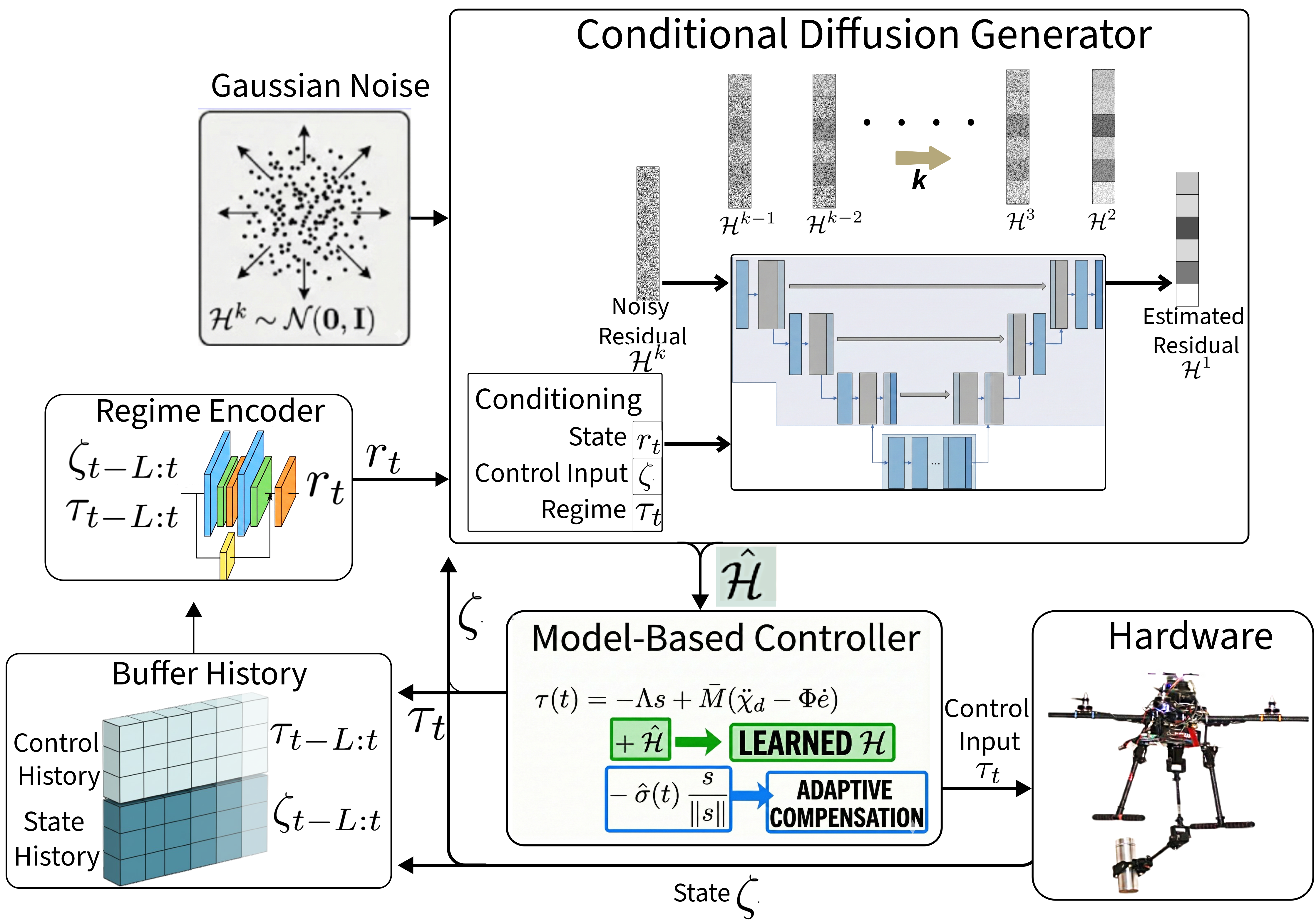}
    \caption{ \small
Closed-loop architecture of the proposed regime-conditioned diffusion framework. 
A rolling history of past states and control inputs are encoded by a TCN to produce the regime descriptor $r_t$. 
The conditional diffusion model uses $(\zeta,\tau,r)$ to generate the residual estimate $\hat{\mathcal{H}}$, which is supplied to the adaptive controller for real-time compensation. 
The applied control input drives the aerial manipulator, and the resulting sensor feedback updates the history buffer, closing the loop.
}
    \label{fig:pipeline}
\end{figure}

% { \color{blue}
% \begin{figure}[!h]
% \centering
% \includegraphics[scale=0.3]{figures/combined_tsne.png}   
% \caption{\small DUMMY: t-SNE plots of $\mathcal{H}$ observed from real-world flight data ({\color{red}red box} in Figure \ref{fig:multimodal}) and $\hat{\mathcal{H}}$ obtained from the baselines and the proposed Diffusion model. While GPT fairly estimates the uncertainties in the dynamics, the proposed diffusion model accurately captures the underlying distribution.}
%     \label{fig:regimes_prediction}
% \vspace{1mm}
% \end{figure}
% Figure \ref{fig:regimes_prediction} illustrates the effectiveness of the proposed regime-conditioned diffusion model in capturing the intricate structure of $\mathcal{H}$.
% }

\subsection{Closed-Loop Deployment with Adaptive Control}
\label{sec:deployment}

Given the residual estimate $\hat{\mathcal{H}}$ produced by the regime-conditioned diffusion model, we now design an adaptive controller to tackle the residual estimation error $\sigma(t)$, characterised by the following standard Lemma.
\begin{lemma}%[\cite{lee2022convergence,de2022convergence}] 
\label{assum1} The residual estimation error $\sigma(t)\triangleq \mathcal{H}(t) - \hat{\mathcal{H}}(t)$ is bounded over compact sets $\zeta \in \mathcal{Z}$, $\tau \in \mathcal{U}$, i.e., $\|\sigma(t)\|\le\sigma_m$ for some unknown constant $\sigma_m$. 
\end{lemma}
%\begin{remark}
Lemma~\ref{assum1} is consistent with convergence guarantees for score-based generative models, which bound the divergence between learned and true data distributions~(\cite{lee2022convergence,de2022convergence}). The goal is to design an adaptive controller without prior knowledge of $\sigma(t)$ and $\sigma_m$.
%\end{remark}

Let us define the tracking error as $e(t) \triangleq \chi(t) - \chi_d(t)$, with $\chi_d$ being the desired coordinates. We introduce an error variable
\begin{equation}
    s(t) = \dot{e}(t) + \Phi e(t),
    \label{sliding_var}
\end{equation}
with $\Phi $ a user-defined positive definite gain matrix. 
Using the nominal inertia matrix $\bar{M}$ and the diffusion-based residual estimate $\hat{\mathcal{H}}$, the control input is designed as 
\begin{subequations}
\label{control_law_adaptive_main}
\begin{align}
    \tau(t)
        &= -\Lambda s
           + \bar{M}(\ddot{\chi}_d - \Phi\dot{e})
           + \hat{\mathcal{H}}
           - \hat{\sigma}(t)\,\frac{s}{\|s\|},
           \label{input_main}\\[1mm]
    \dot{\hat{\sigma}}(t)
        &= \|s\| - \nu\,\hat{\sigma}(t),
        \qquad \hat{\sigma}(0) > 0,
        \label{adaptive_law_main}
\end{align}
\end{subequations}
where positive definite matrix $\Lambda$ and scalar $\nu>0$ are the design parameters; the %$\hat{\mathcal{H}}$ provides feedforward compensation based on the learned residual model, while the 
adaptive gain $\hat{\sigma}(t)$ counteracts the residual estimation errors $\sigma(t)$. %This hybrid structure exploits the predictive capability of the diffusion model and the robustness of the adaptive law. 

\begin{theorem}
\label{theorem:adaptive_control}
Under Lemma~\ref{assum1}, the closed-loop dynamics of \eqref{eq:residual_def} under the control and adaptive laws \eqref{control_law_adaptive_main} remain Uniformly Ultimately Bounded (UUB) (cf. UUB Definition~4.6 in~(\cite{khalil2015nonlinear})).
\end{theorem}
\textit{Proof}. See Appendix.

\begin{algorithm}
  \caption{Proposed Regime-Conditioned Diffusion and Adaptive Control}
  \label{algo:inference}
  \begin{algorithmic}[1]
    \Require Diffusion denoiser $\epsilon_\theta$, regime encoder $f_\phi$,
             noise schedule $\{\Sigma^k\}$, gains $\bar{M}, \Lambda, \Phi, \nu$, window length $L$
    \State \textbf{Initialize:} control input $\tau=0$, adaptive gain $\hat{\sigma}(0)>0$
    \State Obtain initial sensor feedback $\zeta_0$
    \State Initialize buffer $\mathcal{B}  \leftarrow \emptyset$
    \While{task not completed}
        \State Receive desired waypoint $\zeta_d = (\chi_d,\dot{\chi}_d,\ddot{\chi}_d)$
        \State Obtain sensor feedback $\zeta_t$ and update buffer $\mathcal{B}$
        \If{buffer not yet full} \quad $\hat{\mathcal{H}} = 0$
        \Else
            \State Compute regime descriptor $r_t = f_\phi(\mathcal{B})$
            \State Initialize $\mathcal{H}^T \sim \mathcal{N}(0,I)$
            \For{$k=T,\ldots,1$}
               \State {\footnotesize  $(\mu^{k-1},\Sigma^{k-1}) \leftarrow                   \mathrm{Denoise}(\mathcal{H}^{k},\epsilon_\theta(\zeta_t,\tau,r_t,\mathcal{H}^{k},k))$}
                \State $\mathcal{H}^{k-1}\sim\mathcal{N}(\mu^{k-1},\Sigma^{k-1})$
            \EndFor
            \State $\hat{\mathcal{H}} = \mathcal{H}^1$
        \EndIf
        \State Compute $e$, $\dot{e}$ and $s$ using \eqref{sliding_var}
        \State Update $\hat{\sigma}$ using \eqref{adaptive_law_main}
        \State Update control input $\tau$ using \eqref{input_main}
    \EndWhile
  \end{algorithmic}
\end{algorithm}

Algorithm~\ref{algo:inference} and Figure~\ref{fig:pipeline} summarize the closed-loop pipeline: the diffusion model provides the residual estimate $\hat{\mathcal{H}}$, and the adaptive control law \eqref{control_law_adaptive_main} compensates for residual estimation errors and disturbances, ensuring robust trajectory tracking. This hybrid structure exploits the predictive capability of the diffusion model and the robustness of the adaptive law.

\section{Experimental Validation and Analysis}
We evaluate our proposed framework under realistic sensing, actuation, and environmental uncertainties. Here, we assess both (i) model validation: model accuracy in open-loop prediction and (ii) trajectory tracking: control performance in a challenging
payload disturbance scenario. 

\textbf{Hardware Setup:}
For experimental validation, we built an aerial manipulator (cf. Fig. \ref{fig:real_exp_trajectory}) using a Tarot-650 quadrotor frame with SunnySky V4006 brushless motors, a 6S LiPo battery, and 14-inch propellers. A 2R serial-link manipulator (both link lengths $\approx 18$ cm each), actuated by Dynamixel XM430-W210-T motors, is mounted on the quadrotor. The full system weighs approximately $3.0$ kg. A U2D2 Power Hub Board supplies power to the manipulator and gripper. The quadrotor is equipped with a CUAV X7+ flight controller running customized PX4 firmware and an onboard computer, Jetson Orin Nano Super. 
Communication between the Jetson and flight controller uses Micro XRCE-DDS for efficient, low-latency data exchange of PX4 uORB topics. 
Manipulator's joint actuation is handled through the \textit{ros2\_control} framework in current-based torque mode via the Joint Trajectory Controller (JTC).  
State feedback is obtained from an OptiTrack motion capture system (120 fps) fused with onboard IMU data for the quadrotor, while manipulator joint positions and velocities are provided by Dynamixel encoders.  
For inference, a host computer with an NVIDIA RTX 4080 GPU communicates with the onboard Jetson over WiFi. Sensor data are streamed to the host, where the learned dynamics model and control inputs are computed. The resulting body moments and collective thrust commands are transmitted back to the quadrotor at 50 Hz. 

\textbf{Baselines:}
We compare the proposed method against both classical and learning-based alternatives. 
For model validation, we implement a standard SysID pipeline using sparse regression~(\cite{kaiser2018sparse}) and
for tracking, we include a first-principles adaptive sliding mode controller (ASMC)~(\cite{11098573}). 
We further evaluate several established data-driven residual learners: Gaussian Processes~(\cite{torrente2021data}), deep neural networks~(\cite{shi2019neural}), diffusion model without regime conditioning~(\cite{das2025dronediffusion}), and an autoregressive GPT-2 model~(\cite{chen2021decision})\footnote{\url{https://github.com/kzl/decision-transformer}}. 
All learning-based baselines are trained on the same offline dataset to predict $\hat{\mathcal{H}}$ and are deployed using the control law~\eqref{control_law_adaptive_main}. 
All results are reported over $10$ independent trials.

\textbf{Data Collection:}
To collect diverse training data, we use a baseline PID controller to fly randomized trajectories that excite the coupled dynamics of both the aerial platform and the manipulator. Each trajectory includes smooth variations in position, velocity, and joint angles, along with pick–and–drop actions to induce rapid changes in payload. Data are gathered under three payload conditions (0\,g, 200\,g, 400\,g) to generate configuration- and load-dependent variations in inertia and interaction forces. For each condition, $5$ minutes of data are recorded at 100\,Hz, forming a time-indexed dataset  
\(
\mathcal{D}_{\text{raw}}
    = \{(\chi^{(i)}, \dot{\chi}^{(i)}, \tau^{(i)})\}_{i=1}^{N_{raw}} .
\)
Residuals $\mathcal{H}$ are computed using \eqref{eq:residual_def}, and the raw sequences are segmented into fixed-length windows to obtain the training set  
\(
\mathcal{D} = \{\mathcal{S}\}_{i=1}^N.
\)

\begin{figure}[!h]
\centering
\includegraphics[width=0.98\linewidth]{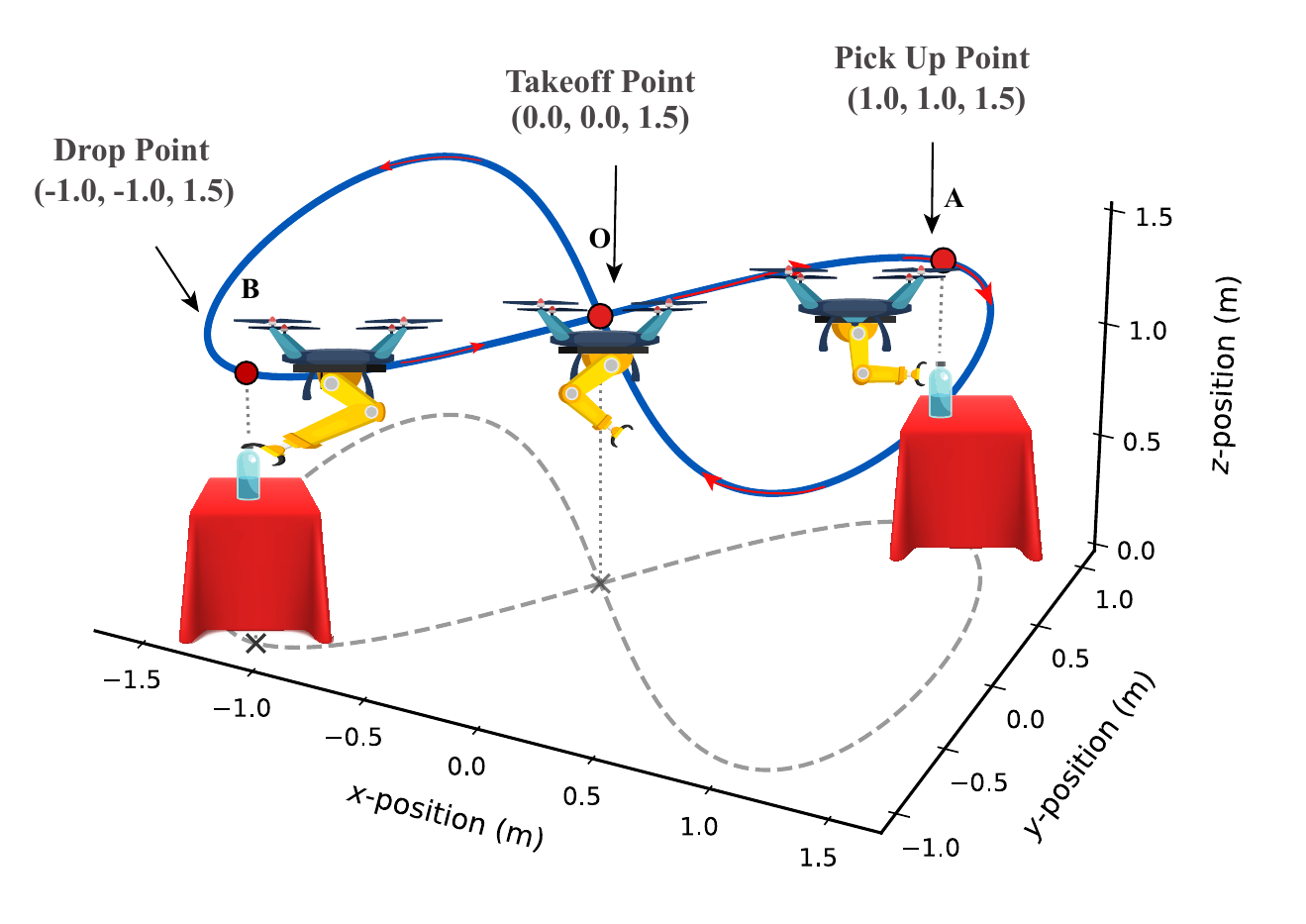}
    \caption{The infinity-shaped trajectory executed during the experiment. The aerial manipulator picks up the payload at point A and releases it at point B before completing the path.}
    \label{fig:exp_trajectory}
\end{figure}

\textbf{Experimental Scenario:}
A pick-and-place task is conducted to evaluate dynamics prediction and control performance (cf. Fig.~\ref{fig:exp_trajectory}). The aerial manipulator transports a payload from Point~A $(1.0,;1.0,;1.5)$ to Point~B $(-1.0,;-1.0,;1.5)$ at average speeds of $0.5$ m/s and $1.0$ m/s. Two payloads of $300$ g and $500$ g ($\approx 45$\% and $50$\% throttle) are used to induce regime shifts due to inertial variation. The trajectory follows an infinity-shaped path in the $XY$ plane with four phases: (i) takeoff from the origin to $z=1.5$ m with the manipulator tucked $(\alpha_1=0^\circ,\alpha_2=90^\circ)$ and transit to A; (ii) grasp at A with manipulator reconfigured to $(50^\circ,50^\circ)$; (iii) transit to B with payload using release configuration $(-50^\circ,-50^\circ)$; (iv) release at B followed by autonomous return and landing. The phases are shown in Fig.~\ref{fig:exp_trajectory}.

\textbf{Implementation:}
We follow the DroneDiffusion (\cite{das2025dronediffusion}) architecture, parameterizing $\epsilon_\theta$ with a temporal U-Net using residual temporal convolution blocks and learned timestep and condition embeddings. The model is trained with Adam ($2{\times}10^{-4}$, batch 256) for 50k steps with $K=20$ diffusion iterations.
The regime descriptor $r_t$ is computed from a sliding history of $L{=}10$ past states using a TCN encoder $f_\phi$ with two dilated 1D convolution layers (kernel size 3, 64 channels), group normalization, and \texttt{tanh} activation, producing a 32-dimensional embedding. Its parameters are trained jointly with the diffusion model. The framework is implemented in JAX with GPU acceleration.

The adaptive controller gains used in all experiments are listed in Table~\ref{tab:controller_params}. %Attitude tracking is implemented using the geometric controller of~\cite{mellinger2011minimum} to calculate 
The desired roll and pitch of the quadrotor are calculated following (\cite{mellinger2011minimum}). 
For all learned models, the input features consist of the current state $\chi_t$, its velocity $\dot{\chi}_t$, and the previous control input $\tau_{t-1}$, since $\tau_t$ is not available at prediction time~(\cite{shi2019neural}).

 \begin{table}[h]
\centering
\caption{Controller Parameter Settings}
\footnotesize
\renewcommand{\arraystretch}{1.2}
\begin{tabular}{l l}
\toprule
$\Phi$ & $\diag\{1.0,\,1.0,\,1.5,\,1.1,\,1.1,\,1.0,\,1.2,\,1.2\}$ \\
$\Lambda$ & $\diag\{2.0,\,2.0,\,3.5,\,1.5,\,1.5,\,1.2,\,3.0,\,3.0\}$ \\
$\bar{M}$ & $\diag\{2,\,2,\,2,\,0.02,\,0.02,\,0.02,\,0.05,\,0.05\}$ \\
$\hat{\sigma}(0)$ & $0.1$ \\
$\nu$ & $2.0$ \\
\bottomrule
\end{tabular}
\label{tab:controller_params}
\end{table}

\subsection{Model Validation}
For predictive accuracy evaluation, the quadrotor tracks a reference trajectory using a vanilla PID controller with a payload of $100$ g is attached to end-effector while $\mathcal{H}$ is computed at each timestep from \eqref{eq:residual_def}. We evaluate the model prediction error of all model-learning baselines—SysID, GP, DNN, GPT-2, Diffusion (without regime conditioning), and our proposed regime-conditioned diffusion model. 
The results are reported in Table~\ref{tab:model_validation}.

\begin{table}[H]
\footnotesize
\renewcommand{\arraystretch}{1.3}
\caption{\small Model prediction performance comparison (RMSE). Lower is better.}
\centering
\scalebox{0.750}{
\begin{tabular}{lccc ccc cc}
\toprule
\multirow{2}{*}{\textbf{Method}} 
    & \multicolumn{3}{c}{\textbf{Position}} 
    & \multicolumn{3}{c}{\textbf{Attitude}}
    & \multicolumn{2}{c}{\textbf{Manipulator}} \\
\cmidrule(r){2-4} \cmidrule(r){5-7} \cmidrule(r){8-9}
    & $\mathcal{H}[1]$ & $\mathcal{H}[2]$ & $\mathcal{H}[3]$
    & $\mathcal{H}[4]$ & $\mathcal{H}[5]$ & $\mathcal{H}[6]$
    & $\mathcal{H}[7]$ & $\mathcal{H}[8]$ \\
\midrule
\textbf{SysID}          
    & 0.125 & 0.133 & 0.179
    & 0.175 & 0.178 & 0.182
    & 0.190 & 0.198 \\
\textbf{DNN}             
    & 0.081 & 0.082 & 0.087
    & 0.118 & 0.131 & 0.149
    & 0.107 & 0.112 \\
\textbf{GP}            
    & 0.076 & 0.075 & 0.077
    & 0.109 & 0.124 & 0.132
    & 0.101 & 0.110 \\
\textbf{GPT}            
    & 0.074 & 0.075 & 0.077
    & 0.103 & 0.121 & 0.131
    & 0.099 & 0.109 \\
\textbf{Diffusion} 
    & 0.064 & 0.067 & 0.068
    & 0.091 & 0.101 & 0.113
    & 0.083 & 0.094 \\
\textbf{Proposed}           
    & \textbf{0.045} & \textbf{0.046} & \textbf{0.049}
    & \textbf{0.077} & \textbf{0.081} & \textbf{0.089}
    & \textbf{0.064} & \textbf{0.071} \\
\bottomrule
\end{tabular}
}
\label{tab:model_validation}

\end{table}

\subsection{Tracking Performance}
We evaluate trajectory tracking performance, as shown in Fig. \ref{fig:real_exp_trajectory}, by applying the control law~\eqref{control_law_adaptive_main} and computing the root-mean-square error (RMSE) of error signal $e(t)$ for each method, as reported in Table~\ref{tab:tracking}.

\begin{table}[H]
\footnotesize
\renewcommand{\arraystretch}{1.3}
\caption{\small Tracking comparison under different payloads and velocities (RMSE in meters)}

\centering
\scalebox{0.95}{
\begin{tabular}{lcccc}
\toprule
\multirow{2}{*}{\textbf{Method}} & \multicolumn{2}{c}{\textbf{300 g}} & \multicolumn{2}{c}{\textbf{500 g}} \\
\cmidrule(r){2-3} \cmidrule(r){4-5}
 & $0.5$ m/s & $1.0$ m/s & $0.5$ m/s & $1.0$ m/s \\
\midrule
\textbf{ASMC}        & 0.138 & 0.194 & 0.216 & 0.299 \\
\textbf{DNN}         & 0.125 & 0.155 & 0.179 & 0.237 \\
\textbf{GP}          & 0.122 & 0.146 & 0.171 & 0.224 \\
\textbf{GPT}         & 0.117 & 0.141 & 0.168 & 0.219 \\
\textbf{Diffusion}   & 0.102 & 0.122 & 0.152 & 0.176 \\
\textbf{Proposed}    & \textbf{0.083} & \textbf{0.102} & \textbf{0.111} & \textbf{0.144} \\
\bottomrule

\end{tabular}
}
\label{tab:tracking}

\end{table}

\subsection{Analysis}
As seen in Tables~\ref{tab:model_validation} and~\ref{tab:tracking}, the performance gap between methods aligns cleanly with their ability to represent the structure of the residual dynamics.
Classical SysID and ASMC rely on fixed parameterized dynamics and therefore produce the largest prediction errors, with tracking performance degrading sharply as speed or payload increases. DNNs reduce this gap, but their deterministic mapping forces them to average residual effects across conditions, leading to underfitting when the arm configuration or payload changes. GPs perform slightly better thanks to their built-in uncertainty modeling, yet their smooth kernel assumptions prevent them from capturing the fast variations introduced by arm motion or payload transitions, causing errors to rise at higher speeds. GPT-based autoregressive models match GP-level prediction accuracy, but their step-wise rollout accumulates errors and does not reliably distinguish residual responses associated with different operating conditions, limiting their closed-loop performance.

Diffusion without regime conditioning performs noticeably better because multi-scale noise training captures the geometry of the residual dynamics and yields accurate multistep predictions, instead of collapsing the residuals into a single averaged estimate. However, since it conditions only on the instantaneous state and input, it cannot distinguish situations where similar measurements arise from different operating conditions, causing its performance to degrade once the manipulator moves or the payload changes.

The proposed regime-conditioned diffusion model provides the largest improvement. By incorporating the latent descriptor $r_t$, the model differentiates residual behaviors arising from distinct configurations, payload masses, and recent motions, yielding a clear reduction in prediction error over the unconditioned diffusion baseline and the lowest tracking error across all payloads and speeds. Even for the 500 g case, outside the training distribution, the method maintains the smallest RMSE among all evaluated models.

\section{Conclusion}
We presented a regime-conditioned diffusion framework for learning the residual dynamics of aerial manipulators.
By combining a conditional diffusion model with a compact temporal encoder, the method captures configuration-, payload-, and motion-dependent effects that cannot be represented by standard deterministic or single-regime models.
Integrated with an adaptive controller, the learned residuals provide reliable real-time compensation under rapid configuration changes and out-of-distribution payloads.
Real-world experiments demonstrate substantial gains in both prediction accuracy and closed-loop tracking performance over classical and data-driven baselines.
These results highlight the value of distribution-aware residual modeling for robust aerial manipulation.

\bibliographystyle{IEEEtran}
\bibliography{our_bib} 

\newpage
\appendix

\section{Proof of Theorem 2.2} \label{adaptive_control_design}
Multiplying $\dot{s}$ from~\eqref{sliding_var} by $\bar{M}$ and using the translational dynamics~\eqref{eq:residual_def} yields
\begin{equation}
    \bar{M}\dot{s}
      = \bar{M}(\ddot{\chi} - \ddot{\chi}_d + \Phi\dot{e})
      = \tau - \mathcal{H} - \bar{M}(\ddot{\chi}_d - \Phi\dot{e}).
    \label{dot_s}
\end{equation}
Substituting the control input~\eqref{input_main} into~\eqref{dot_s} gives
\begin{align}
    \bar{M}\dot{s}
        &= -\Lambda s + \hat{\mathcal{H}} - \mathcal{H}
           - \hat{\sigma}(t)\frac{s}{\|s\|} \nonumber\\
        &= -\Lambda s + \sigma
           - \hat{\sigma}(t)\frac{s}{\|s\|},
    \label{ms_dot}
\end{align}
where $\sigma \triangleq \mathcal{H} - \hat{\mathcal{H}}$ is the residual estimation error.

\textbf{\textit{Proof.}}
Consider the Lyapunov function candidate
\begin{equation}
    \mathcal{V}
      = \frac{1}{2}s^{\top}\bar{M}s
        + \frac{1}{2}(\hat{\sigma} - \sigma_m)^2.
    \label{lyapunov_fcn}
\end{equation}
Differentiating~\eqref{lyapunov_fcn} and using~\eqref{ms_dot} gives
\begin{align}
\dot{\mathcal{V}}
    &= s^{\top}\bar{M}\dot{s}
       + (\hat{\sigma} - \sigma_m)\dot{\hat{\sigma}} \nonumber\\
    &= s^{\top}
         \bigl(
             -\Lambda s + \sigma
             - \hat{\sigma}\frac{s}{\|s\|}
         \bigr)
       + (\hat{\sigma} - \sigma_m)\dot{\hat{\sigma}} \nonumber\\
    &\le -\lambda_{\min}(\Lambda)\|s\|^2
          + \|s\|\|\sigma\|
          - \hat{\sigma}\|s\|
          + (\hat{\sigma} - \sigma_m)\dot{\hat{\sigma}}.
    \label{v_dot_intermediate}
\end{align}

The adaptive law~\eqref{adaptive_law_main} yields
\begin{equation}
(\hat{\sigma} - \sigma_m) \dot{\hat{\sigma}} = \norm{s}(\hat{\sigma} - \sigma_m) + \nu \hat{\sigma} \sigma_m -\nu \hat{\sigma}^2.
    \label{adaptive_term}
\end{equation}

Using Lemma~\ref{assum1} i.e. $\|\sigma\|\le\sigma_m$, and substituting~\eqref{adaptive_term} into~\eqref{v_dot_intermediate}, we obtain
\begin{align}
\dot{\mathcal{V}} 
&\leq - \lambda_{\min}(\Lambda) \norm{s}^{2} - (\hat{\sigma} - \sigma_m)(\norm{s}  - \dot{\hat{\sigma}} ) \nonumber \\
&= - \lambda_{\min}(\Lambda) \norm{s}^{2} + (\nu \hat{\sigma} \sigma_m -\nu \hat{\sigma}^2 ) \nonumber \\
&\leq - \lambda_{\min}(\Lambda) \norm{s}^{2} - \frac{\nu}{2}  ((\hat{\sigma} - \sigma_m)^2 - {\sigma_m}^2 ). \label{v_dot_clean}
\end{align}
The definition of $\mathcal{V}$ yields %Using the bound from~\eqref{lyapunov_fcn},
\begin{align} \label{lyap_bound}
    \mathcal{V} \leq \frac{1}{2} \bar{M}||s||^2 + \frac{1}{2} (\hat{\sigma} - \sigma_m)^2. 
\end{align}

Substituting (\ref{lyap_bound}) into (\ref{v_dot_clean}), $\dot{\mathcal{V}}$ is simplified to
\begin{align}
 \dot{\mathcal{V}}
   \le -\rho\,\mathcal{V} + \delta,   
\end{align}
where
$
\rho
   \triangleq
   \frac{\min\{\lambda_{\min}(\Lambda),\,\nu/2\}}
        {\max\{\lambda_{\min}{M}/2,\, 1/2\}}
   > 0,
\qquad
\delta \triangleq \frac{\nu}{2}\sigma_m^2.
$

Standard comparison arguments show that for any $0<\kappa<\rho$,
\[
\dot{\mathcal{V}}
    \le -\kappa\mathcal{V}
       -(\rho-\kappa)\mathcal{V} + \delta.
\]
Define $\mathcal{B}=\delta/(\rho-\kappa)$. Then whenever $\mathcal{V}(t)\ge\mathcal{B}$,
\[
\dot{\mathcal{V}}(t) \le -\kappa\mathcal{V}(t),
\]
which implies
\[
\mathcal{V}(t)
    \le \max\{\mathcal{V}(0),\mathcal{B}\},\qquad \forall t\ge 0.
\]
thus, all closed-loop trajectories remain UUB.

\begin{remark}
To ensure continuity of the control input, the term $s/\|s\|$ in \eqref{input_main} is
commonly replaced by the smooth approximation
$\frac{s}{\sqrt{\|s\|^2+\varpi}}$ with $\varpi>0$. This modification preserves the UUB
property. In our experiments we use $\varpi=0.1$. % to reduce chattering.
\end{remark}
                                                                         % in the appendices.
\end{document}